\documentclass[10pt,leqno]{amsart}
\usepackage{graphicx}
\usepackage{indentfirst,csquotes}

\usepackage{amssymb,amsthm,amsmath}
\usepackage{xcolor,paralist,hyperref,titlesec,fancyhdr,etoolbox}

\titleformat{\section}{\normalfont\large\bfseries}{\thesection.}{0.6em}{}
\titlespacing*{\section}{0pt}{2.5ex}{1.2ex}
\titleformat{\subsection}{\normalfont\normalsize\bfseries}{\thesubsection.}{0.5em}{}
\titlespacing*{\subsection}{0pt}{1.8ex}{0.8ex}
\makeatletter\providecommand{\@secnumpunct}{.}\makeatother

\hypersetup{ colorlinks=true, linkcolor=black, filecolor=black, urlcolor=black, citecolor=blue }

\usepackage{booktabs,multirow,url}
\usepackage{algorithm}
\usepackage{algpseudocode}
\graphicspath{{figures/}}

\newcommand{\pmci}[1]{\,{\scriptsize$\pm$#1}}
\newcommand{\figorbox}[2]{%
  \IfFileExists{figures/#1.pdf}{\includegraphics[width=#2]{#1.pdf}}{%
  \IfFileExists{figures/#1.png}{\includegraphics[width=#2]{#1.png}}{%
  \fbox{\parbox[c][4cm][c]{\dimexpr#2-2\fboxsep-2\fboxrule\relax}{\centering Missing figure: \texttt{figures/\detokenize{#1}.pdf}}}}}}

\begin{document}
\title{Component Pooling Trained Across Orientations for Few-Shot Remote Sensing Scene Classification}

\date{}
\maketitle

\begin{center}
  \small
  \textbf{Bingchen Yan}\\
  School of Computer Science and Information Security,\\
  Guilin University of Electronic Technology, Guilin, 541004, China\\
  \texttt{yanbingchen132@126.com}\\[6pt]

  \textbf{Wenheng Guo}\\
  School of Information Engineering,\\
  Institute of Disaster Prevention, Langfang, 065201, China\\

  \textbf{Yuye Zhou}$^{*}$\\
  School of Computing \& Technology,\\
  Asia Pacific University of Technology \& Innovation, Kuala Lumpur, 57000, Malaysia\\

  $^{*}$Corresponding author
\end{center}

\date{\today}

\begin{abstract}
Few-shot remote sensing scene classification (FSRSSC) aims to recognize novel scene categories from only a few labeled images. Most FSRSSC methods summarize the feature map of an image by global average pooling, which treats the few objects that define an aerial scene and the large background around them alike, and they rarely consider that aerial images have no canonical orientation. To address these issues, we propose RoCoP, a parameter-free component pooling network trained across orientations. RoCoP consists of a feature extractor, a non-negative decomposition module (NDM) and a component scoring module (CSM). NDM factorizes the local descriptors of each image into a few additive components with a deterministic initialization and a single multiplicative update, so that the positions of one object are grouped together. CSM weights each component by its activation magnitude and converts the weights into a soft mask for pooling. Because both modules are invariant to permutations of feature positions, the only rotation-sensitive part of the pipeline is the backbone, which we train to classify randomly rotated views of every image. Experiments on NWPU-RESISC45, AID and UC Merced show that component pooling and rotated-view training are complementary: together they improve a strong transfer-learning baseline by about 2.5 points in 5-way 1-shot accuracy on NWPU-RESISC45 and AID, without adding parameters at inference. Under an identical protocol, RoCoP outperforms re-implemented metric-based and task-adaptive methods, and NDM and CSM also improve existing methods when plugged into them. We further identify the regime in which the method does not apply, namely very small base sets.
\end{abstract}
\maketitle

\section{Introduction}
Remote sensing scene classification assigns a semantic label, such as airport, farmland or residential area, to an aerial or satellite image, and supports land-use mapping, urban planning and environmental monitoring~\cite{cheng2017nwpu,xia2017aid}. Deep convolutional networks have made this task largely solved when thousands of labeled images per class are available~\cite{penatti2015deep,cheng2017nwpu}. Labeling remote sensing images, however, requires expert knowledge, and new categories appear as applications change. Few-shot remote sensing scene classification (FSRSSC) therefore asks a model trained on a set of base classes to recognize novel classes from one or five labeled images per class~\cite{qiu2024fsrssc,dlamatchnet,spnet}.

Existing few-shot methods follow three main paradigms. Optimization-based meta-learning learns an initialization that adapts quickly to a new task~\cite{finn2017maml}. Metric-based methods learn an embedding in which a query is classified by its distance to the labeled support samples~\cite{vinyals2016matching,snell2017proto,sung2018relation}. Transfer-learning methods train an embedding on all base classes and fit a simple classifier on the support set of each task, and they have proved to be strong baselines~\cite{chen2019closer,tian2020rfs,wang2019simpleshot}. In remote sensing, these ideas have been extended with attention~\cite{dlamatchnet,xu2024acl}, prototype calibration~\cite{spnet,wang2023tdnet,zhu2025diffpr}, kernel and subspace metrics~\cite{cui2022mkn,qin2024dusn}, contrastive and self-supervised learning~\cite{li2022sclmlnet,zeng2022tscl,ma2023mpcl,rgscs2026}, relation modeling~\cite{tian2024hirenet,zhang2022sgmnet}, multi-scale feature fusion~\cite{wu2026came,yang2026mmff}, and, more recently, vision--language and pretrained models~\cite{zheng2026t2net,gu2026s3opt}.

Two properties of aerial scenes are, however, rarely addressed together. First, the evidence for a scene class is usually carried by a few ground objects, such as the runways of an airport or the courts of a sports field, while bare land, vegetation or water fills most of the image. Most FSRSSC pipelines nevertheless summarize the feature map by global average pooling (GAP), which weights object and background positions equally. Methods that keep local descriptors~\cite{li2019dn4,zhang2020deepemd,wertheimer2021frn,qin2025ldrn,wang2025classaware} or weight them with attention~\cite{xu2024acl,jia2025pfa} judge each position independently or learn attention parameters on the base classes, so the positions of one object may receive inconsistent weights on novel classes. Second, aerial images are taken from above and have no canonical orientation. Rotation is usually handled by data augmentation or by self-supervised rotation prediction~\cite{gidaris2019boost,ma2023mpcl}, independently of how the features are pooled.

In this paper, we propose RoCoP, which pools features over components rather than positions and trains the backbone across orientations. A non-negative decomposition module (NDM) groups positions whose channel responses are alike into a few additive components by non-negative matrix factorization (NMF)~\cite{lee1999nmf}, and a component scoring module (CSM) weights each component as a unit, so that all positions of one object share one weight. Neither module has learnable parameters. We show that both modules are invariant to permutations of feature positions, so the only rotation-sensitive part of the pipeline is the backbone; RoCoP therefore trains the backbone to classify randomly rotated views of every image, instead of adding constraints to the pooling. Our experiments show that component pooling alone brings a modest gain, that the gain becomes much larger when the backbone is trained with rotated views, and that two natural extensions, an explicit mask-consistency loss and a task-level calibration of the component scores, do not help.

The main contributions of this paper are summarized as follows.
\begin{enumerate}
\item We propose a parameter-free component pooling for FSRSSC, composed of NDM and CSM. A deterministic NNDSVD initialization and a single multiplicative update make the decomposition reproducible for every image, cheap and differentiable, so it can be trained end to end while inference remains inductive.
\item We show that component pooling and rotated-view training are complementary. The analysis is supported by paired comparisons on identical test episodes and by the stability of the pooling masks under rotation, and it shows that the plain rotated-view objective is preferable to explicit consistency constraints.
\item We conduct extensive experiments on three benchmarks, including same-protocol re-implementations of representative methods, plug-in studies, and analyses of the classifier, the calibration and the number of components. RoCoP outperforms the re-implemented methods, improves other methods as a plug-in module, and its limitation on very small base sets is identified.
\end{enumerate}

The remainder of this paper is organized as follows. Section~\ref{sec:related} reviews related work. Section~\ref{sec:method} describes RoCoP. Section~\ref{sec:exp} reports the experiments, Section~\ref{sec:discussion} discusses the limitations, and Section~\ref{sec:conclusion} concludes the paper.

\section{Related Work}\label{sec:related}
\subsection{Few-Shot Learning}
Optimization-based methods, represented by MAML~\cite{finn2017maml}, learn model parameters that can be adapted to a new task with a few gradient steps. Metric-based methods learn a feature space and a comparison rule: matching networks~\cite{vinyals2016matching} use attention over the support set, prototypical networks~\cite{snell2017proto} compare queries with class means, and relation networks~\cite{sung2018relation} learn the metric itself. Later work matches local descriptors~\cite{li2019dn4,zhang2020deepemd,wertheimer2021frn}, models second-order statistics~\cite{xie2022deepbdc}, or adapts the embedding to each task~\cite{ye2020feat,hou2019can}. A line of transfer-learning studies showed that a backbone trained with standard cross-entropy on all base classes, followed by a linear or nearest-centroid classifier, is competitive with these methods~\cite{chen2019closer,tian2020rfs,wang2019simpleshot}. We adopt this transfer-learning setting and change only how the feature map is pooled and how the backbone is trained.

\subsection{Few-Shot Remote Sensing Scene Classification}
Early FSRSSC methods adapted metric learning to aerial images. RS-MetaNet~\cite{rsmetanet} learns a metric over tasks, DLA-MatchNet~\cite{dlamatchnet} adds discriminative attention to a matching network, and SPNet~\cite{spnet} calibrates prototypes with self- and inter-calibration. Later methods enlarge the margin between confusing classes with meta-kernels~\cite{cui2022mkn} or subspaces~\cite{qin2024dusn}, rectify prototypes with transductive information~\cite{wang2023tdnet} or diffusion models~\cite{zhu2025diffpr}, model relations between samples and scenes~\cite{tian2024hirenet,zhang2022sgmnet}, and fuse multi-scale features~\cite{wu2026came,yang2026mmff}. Self-supervised and contrastive objectives are widely used to learn more transferable features~\cite{li2022sclmlnet,zeng2022tscl,ma2023mpcl,ji2023dcn,li2023mes2l,dong2025dbarmcl}; RGSCS~\cite{rgscs2026} enforces the consistency of reconstruction coefficients across 90$^\circ$ rotations. Recent methods also exploit pretrained convolutional networks~\cite{gu2026s3opt} and vision--language models~\cite{liu2024remoteclip,zheng2026t2net}, which are not directly comparable with models trained from scratch on the base classes. A recent survey~\cite{qiu2024fsrssc} reviews these methods and benchmarks. Unlike them, RoCoP keeps the standard transfer-learning pipeline and changes only the pooling and the training views, so it can be combined with many of these methods.

\subsection{Feature Weighting and Pooling}
GAP is the default way to summarize a feature map. Generalized-mean pooling~\cite{radenovic2019gem} and bilinear or second-order pooling~\cite{lin2015bilinear,xie2022deepbdc,deng2024msop} change the statistic that is pooled, while channel and spatial attention~\cite{hu2018senet,woo2018cbam} learn which features to emphasize. In FSRSSC, attention modules highlight object regions~\cite{dlamatchnet,xu2024acl}, a parameter-free attention selects discriminative regions for matching~\cite{jia2025pfa}, and local-descriptor methods rectify or reweight descriptors~\cite{qin2025ldrn,wang2025classaware}. These methods weight each position independently or learn their weights on base classes. NDM instead groups positions first, using NMF, which yields parts-based representations~\cite{lee1999nmf,lee2001nmfalg}. NMF of deep features has been used to discover object parts in trained networks~\cite{collins2018dff} and to learn hierarchical representations~\cite{trigeorgis2014deepsemi}; we use it as a differentiable pooling operator at both training and test time.

\subsection{Rotation in Remote Sensing}
Rotation-invariant and rotation-equivariant networks build orientation into the architecture through rotation-invariant layers~\cite{cheng2016ricnn}, group convolutions~\cite{cohen2016gcnn,weiler2019e2cnn} or rotation-equivariant detectors~\cite{redet}. Self-supervised rotation prediction~\cite{gidaris2018rot} has been used as an auxiliary task for few-shot learning~\cite{gidaris2019boost} and for FSRSSC~\cite{ma2023mpcl}. We use the simplest form, classifying rotated copies of each training image with the original label, and show that it changes what component pooling learns.

\section{Method}\label{sec:method}
\begin{figure*}[t]
\centering
\figorbox{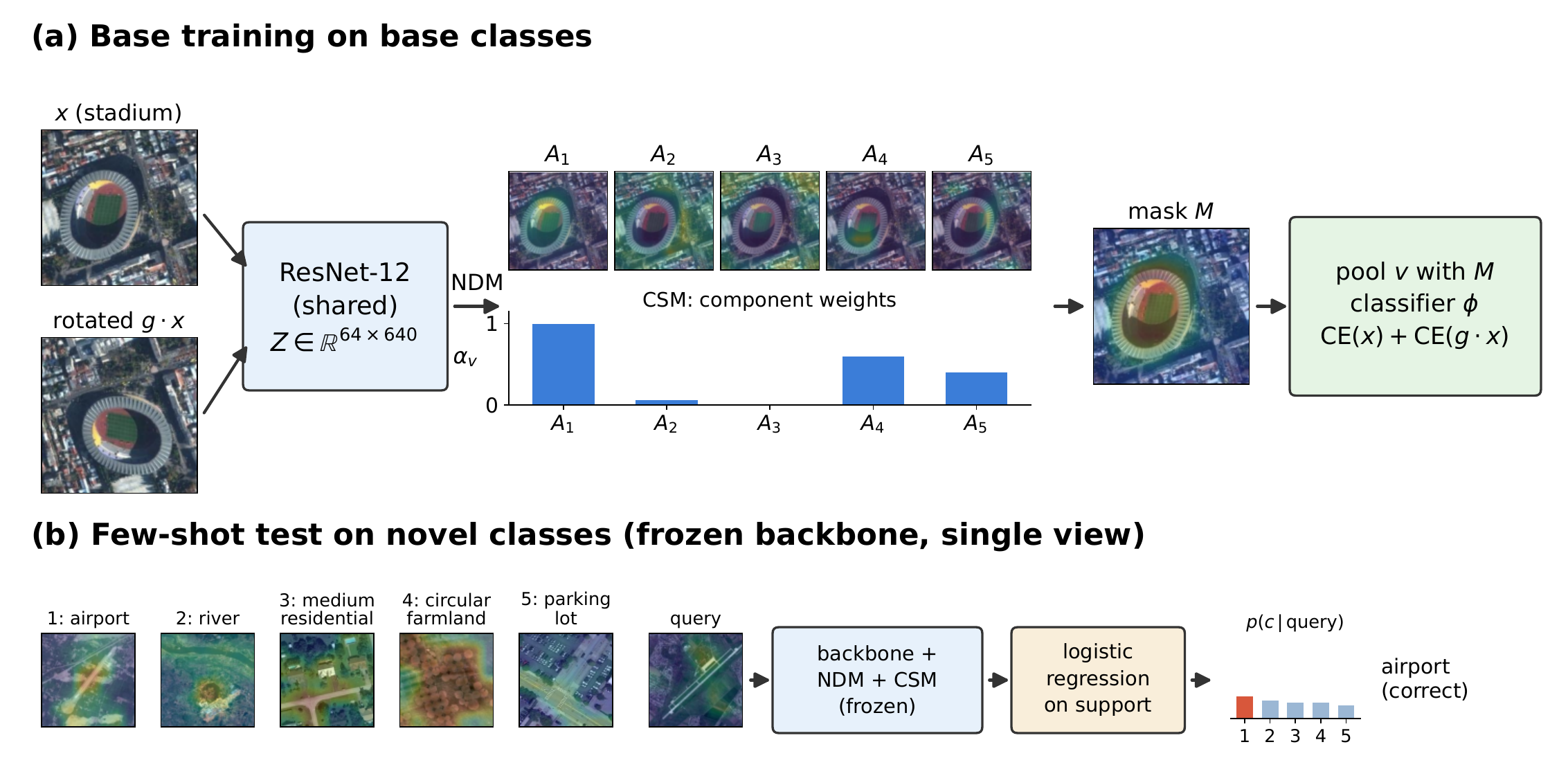}{\textwidth}
\caption{Overview of RoCoP, drawn with real NWPU-RESISC45 images and the outputs of the trained model. (a) Base training on base classes: an image and a rotated copy pass through a shared ResNet-12; NDM factorizes the $8\times8\times640$ feature map into $r=5$ components ($A_1$--$A_5$, overlaid on the image), CSM scores each component by its mean descriptor norm and scales the scores within the image, and the resulting mask $M$ pools the descriptors for a linear classifier trained on both views. (b) Few-shot test on novel classes (5-way 1-shot shown): the frozen backbone, NDM and CSM encode the support images and the query (masks overlaid), and a logistic regression trained on the support vectors predicts the class of the query (bars: class probabilities; red: prediction). No parameters are added at test time.}
\label{fig:overview}
\end{figure*}

\subsection{Problem Setting and Overview}
The base classes $\mathcal{C}_b$ and the novel classes $\mathcal{C}_n$ are disjoint. A ResNet-12 backbone~\cite{oreshkin2018tadam} $f$ is trained on $\mathcal{C}_b$. At test time $f$ is frozen and each $N$-way $K$-shot episode provides a support set $\mathcal{S}=\{(x_i,y_i)\}_{i=1}^{NK}$ and query images from the same $N$ novel classes. We use the inductive setting: queries are classified independently and no statistic is shared across the query set.

RoCoP has three parts (Fig.~\ref{fig:overview}). (i) The backbone maps an image to non-negative local descriptors; all activations of $f$ are ReLU, and a $128\times128$ input gives an $8\times8\times640$ map. (ii) NDM groups the descriptors into $r$ additive components with soft assignment maps (Sec.~\ref{sec:ndm}). (iii) CSM scores the components and turns them into a soft mask (Sec.~\ref{sec:csm}). The mask pools the descriptors, and the pooled vectors are classified (Sec.~\ref{sec:cls}). The backbone is trained with rotated views (Sec.~\ref{sec:rot}).

\subsection{Non-negative Decomposition Module (NDM)}\label{sec:ndm}
\textbf{Motivation.} An aerial scene is evidence from a few objects laid on large backgrounds. The descriptors of one object are similar to each other across the positions it covers, and differ from background descriptors. GAP ignores this structure. Position-wise weighting uses it only partly: each position is judged alone, so an object can receive uneven weights across its own positions. NDM first groups positions whose channel responses are alike into a few additive components, so that CSM can weight each component as a unit and all positions of one object share one weight. Because the descriptors after ReLU are non-negative, NMF~\cite{lee1999nmf} is a natural grouping: it explains each descriptor as an additive mixture of a few parts.

\textbf{Notation.} Reshape the last feature map of one image into $Z=[z_1,\dots,z_L]^\top\in\mathbb{R}_{\ge0}^{L\times C}$ with $L=64$ positions and $C=640$ channels. NDM approximates
\begin{equation}
\min_{W\ge0,\,H\ge0}\ \tfrac12\lVert Z-WH\rVert_F^2
\end{equation}
with $W\in\mathbb{R}_{\ge0}^{L\times r}$ (how much of each component every position contains) and $H\in\mathbb{R}_{\ge0}^{r\times C}$ (the channel profile of each component), $r=5$. The factors are recomputed for every support and query image, and NDM stores no weights. The construction below approaches this objective rather than solving it exactly.

\textbf{Step 1: deterministic initialization.} Let $Z\approx\sum_{j=1}^{r}\sigma_j u_j v_j^\top$ be the rank-$r$ truncated SVD. NNDSVD~\cite{boutsidis2008nndsvd} builds a non-negative start from it. The leading pair is non-negative up to sign, so
\begin{equation}
W_0[:,1]=\sqrt{\sigma_1}\,|u_1|,\qquad H_0[1,:]=\sqrt{\sigma_1}\,|v_1|^\top .
\end{equation}
For $j\ge2$, split $u_j=u_j^+-u_j^-$ and $v_j=v_j^+-v_j^-$ into positive and negative parts, compute $\mu^\pm=\lVert u_j^\pm\rVert\,\lVert v_j^\pm\rVert$, keep the sign pair with the larger $\mu$, and set
\begin{equation}
W_0[:,j]=\sqrt{\sigma_j\mu}\,\frac{u_j^{s}}{\lVert u_j^{s}\rVert},\qquad H_0[j,:]=\sqrt{\sigma_j\mu}\,\frac{(v_j^{s})^\top}{\lVert v_j^{s}\rVert},
\end{equation}
where $s\in\{+,-\}$ is the kept sign and $\mu=\mu^{s}$. Entries are clamped at a small $\epsilon$ so that the multiplicative update below can move them. This start uses neither labels nor random numbers: the same image always yields the same components.

\textbf{Step 2: one multiplicative update.} One Lee--Seung multiplicative update~\cite{lee2001nmfalg} refines the start, first $H$ and then $W$ with the new $H$:
\begin{equation}
H \leftarrow H_0 \odot \frac{W_0^{\top} Z}{W_0^{\top} W_0 H_0+\varepsilon},\qquad
W \leftarrow W_0 \odot \frac{Z H^{\top}}{W_0 H H^{\top}+\varepsilon},
\end{equation}
with element-wise products and quotients and $\varepsilon=10^{-8}$. All terms are non-negative, so no projection is needed.

\textbf{Step 3: assignment maps.} Row-normalizing $W$ gives each position a soft assignment over the components,
\begin{equation}
A_{lv}=\frac{W_{lv}}{\sum_{u=1}^{r} W_{lu}+\varepsilon},\qquad \sum_v A_{lv}\le 1,
\end{equation}
so positions with similar channel responses are assigned to the same component.

\textbf{Gradients.} NNDSVD is treated as a constant in back-propagation; the loss is differentiated through the update with respect to $Z$ and reaches the backbone, so the graph stays shallow and no gradient passes through the SVD.

\subsection{Component Scoring Module (CSM)}\label{sec:csm}
NDM builds the components; CSM estimates their relative importance.

\textbf{Activation-magnitude score.} The $\ell_2$ norm of a descriptor summarizes its activation across channels. CSM scores component $v$ by the assignment-weighted mean descriptor norm,
\begin{equation}\label{eq:score}
s_v = \frac{\sum_l A_{lv}\,\lVert z_l \rVert_2}{\sum_l A_{lv}+\varepsilon}.
\end{equation}
The score needs no labels, so the same expression applies to training, support and query images. The intent is that components with strong, specific responses receive high weights and flat background receives low weights.

\textbf{Per-image calibration.} The scores are min--max scaled within the image,
\begin{equation}\label{eq:range}
\alpha_v=\frac{s_v-\min_u s_u}{\max_u s_u-\min_u s_u},
\end{equation}
with $\alpha_v=\frac12$ for all $v$ when $\max_u s_u=\min_u s_u$. The same rule is used in training and at test time, and the weights of an image depend on that image only, so inference is inductive.

\textbf{Soft mask.} Each position inherits the weights of its components,
\begin{equation}
M_l=\sum_{v=1}^{r}\alpha_v A_{lv}\in[0,1],
\end{equation}
so several high-weight components can contribute to one continuous mask instead of a hard component selection.

\textbf{Task-level variant.} A natural alternative replaces the per-image extrema in Eq.~\eqref{eq:range} by the extrema over all components of all support images, $s_{\min}=\min_{i\in\mathcal{S},v}s^{(i)}_v$ and $s_{\max}=\max_{i\in\mathcal{S},v}s^{(i)}_v$, and clips query weights to $[0,1]$. All images of an episode then share one task-dependent scale, and inference stays inductive because the range uses the support set only. Since a base-training batch has no support set, this variant can only be applied at test time. We evaluate it in Sec.~\ref{sec:calib}; it does not improve accuracy, so RoCoP uses per-image calibration.

\subsection{Component Pooling and Classification}\label{sec:cls}
The mask pools the descriptors,
\begin{equation}\label{eq:pool}
v=\frac{\sum_l M_l z_l}{\max\big(\sum_l M_l,\ \rho_0L\big)},
\end{equation}
with $\rho_0=0.05$; an image whose mask is all zero falls back to uniform pooling. During base training, $v$ is fed to a linear classifier $\phi$ over $\mathcal{C}_b$, which is discarded at test time. For each test episode, support and query vectors are $\ell_2$-normalized and a logistic regression ($C=5$) trained on the support vectors classifies the queries~\cite{tian2020rfs,rgscs2026}. As a second classifier we use mask-weighted prototypes, which pool all $K$ support images of class $c$ jointly, $p_c=\sum_{k}\sum_l M^{(c,k)}_l z^{(c,k)}_l/\max(\sum_k\sum_l M^{(c,k)}_l,\rho_0KL)$, and assign the query to $\arg\max_c\cos(v,p_c)$. Algorithm~\ref{alg:rocop} summarizes the pooling of one image.

\begin{algorithm}[t]
\caption{RoCoP pooling of one image (training and test)}\label{alg:rocop}
\begin{algorithmic}[1]
\Require descriptors $Z=f(x)\in\mathbb{R}_{\ge0}^{L\times C}$, rank $r$, floor $\rho_0$
\State $(W_0,H_0)\gets\textsc{NNDSVD}(Z,r)$ \Comment{NDM, no gradient}
\State $H\gets H_0\odot (W_0^\top Z)\oslash(W_0^\top W_0H_0)$
\State $W\gets W_0\odot (ZH^\top)\oslash(W_0HH^\top)$
\State $A_{lv}\gets W_{lv}/\sum_u W_{lu}$
\State $s_v\gets \sum_l A_{lv}\lVert z_l\rVert/\sum_l A_{lv}$ \Comment{CSM}
\State $\alpha\gets\operatorname{minmax}(s)$;\ \ $M_l\gets\sum_v\alpha_vA_{lv}$
\State \Return $v=\sum_l M_l z_l/\max(\sum_l M_l,\rho_0L)$
\end{algorithmic}
\end{algorithm}

\textbf{Properties.} Two properties follow from the definitions and guide the training design.

\emph{(P1) RoCoP contains GAP.} With $r=1$, every position belongs to the single component ($A_{l1}=1$), the score range is zero, so $\alpha_1=\frac12$, $M_l=\frac12$ for all $l$, and $v=\frac1L\sum_l z_l$, which is GAP. Larger $r$ lets the pooling depart from uniform weighting only where the descriptors form distinct groups.

\emph{(P2) NDM and CSM are invariant to permuting positions.} Let $P$ be an $L\times L$ permutation matrix. The left singular vectors of $PZ$ are $Pu_j$ and the right ones are unchanged, and NNDSVD only takes element-wise positive parts and norms, so it returns $(PW_0,H_0)$ (assuming distinct singular values). In the update, $(PW_0)^\top PZ=W_0^\top Z$ leaves $H$ unchanged and $W$ becomes $PW$. Hence $A$ and $M$ are permuted, the scores $s_v$ are unchanged, and $v(PZ)=v(Z)$. A rotation of a square feature map by a multiple of 90$^\circ$ is such a permutation. NDM and CSM are therefore exactly rotation-consistent; any change of the output under image rotation comes from the backbone, because $f(gx)$ is not the rotated $f(x)$. This motivates training the backbone with rotated views (Sec.~\ref{sec:rot}) rather than adding constraints to the pooling.

\textbf{Design choices.} (i) A deterministic start makes the encoding of an image independent of any random state, which is required when the same image may serve as support in one episode and as query in another. (ii) One update keeps the cost low and the gradient path short; we do not tune the number of updates in this paper. (iii) Scores are scaled within each image, identically in training and test, so no statistic is shared across images.

\textbf{Position-wise control.} Replacing the components by the positions themselves ($A=I$, $r=L$), i.e.\ $M_l=\operatorname{minmax}(\lVert z_l\rVert)$, gives a pooling with the same magnitude cue but no grouping. It is the opposite extreme of (P1) and the main control for the grouping step.

\subsection{Rotated-View Training}\label{sec:rot}
By (P2), the remaining source of rotation sensitivity is the backbone. For each training image $x$ with label $y$, a rotation $g$ is sampled uniformly from $\{0^\circ,90^\circ,180^\circ,270^\circ\}$ and both the original and the rotated view are classified with the same label:
\begin{equation}
\mathcal{L} = \tfrac{1}{2}\Big[\mathrm{CE}\big(\phi(v(x)),\,y\big) + \mathrm{CE}\big(\phi(v(g x)),\,y\big)\Big].
\end{equation}
No rotation is applied at test time. We also tested an explicit mask-consistency term that pulls $M(gx)$ toward the rotated $M(x)$ with weight 0.1; it lowered accuracy (Sec.~\ref{sec:ablation}) and is not part of RoCoP.

\subsection{Complexity}
Per image, the multiplicative update costs $O(LCr+(L+C)r^2)$ and the truncated SVD for NNDSVD $O(LC\min(L,C))$. CSM needs $O(LC+Lr)$ for the scores and $O(Lr)$ for the mask. NDM and CSM add no parameters: all models have 12.42\,M parameters in the backbone. Measured on one NVIDIA RTX 4090 with batch 16 at $128\times128$ (compute only), a training step takes 0.028\,s with GAP and 0.053\,s with NDM, and 0.049\,s and 0.106\,s with rotated views, which double the forward passes. Inference uses one view: encoding and pooling a 5-way 5-shot episode (25 support and 75 query images) takes 44\,ms with GAP and 67\,ms with RoCoP. 


\section{Experiments}\label{sec:exp}
\subsection{Datasets}
We evaluate RoCoP on three widely used benchmarks.

\textbf{NWPU-RESISC45}~\cite{cheng2017nwpu} contains 31\,500 images of $256\times256$ pixels from 45 scene classes, with 700 images per class and spatial resolutions from 0.2 to 30\,m. It is split into 25 base, 10 validation and 10 novel classes.

\textbf{AID}~\cite{xia2017aid} contains 10\,000 images of $600\times600$ pixels from 30 classes, with 220 to 420 images per class. It is split into 16 base, 7 validation and 7 novel classes.

\textbf{UC Merced}~\cite{yang2010ucm} contains 2\,100 images of $256\times256$ pixels from 21 land-use classes, with 100 images per class. It is split into 10 base, 5 validation and 6 novel classes, so that only 1\,000 images are available for base training.

We use the class splits of~\cite{rgscs2026} and remove exact duplicate images (8 in NWPU-RESISC45). All images are resized to $128\times128$.

\subsection{Implementation Details}
The backbone is a ResNet-12~\cite{oreshkin2018tadam,lee2019metaopt} with ReLU activations, trained from scratch on the base classes for 100 epochs with SGD (momentum 0.9, weight decay $5\times10^{-4}$, batch size 16). The learning rate is 0.05 and is divided by 10 at epochs 60 and 80. Data augmentation consists of random resized cropping (scale 0.5--1), horizontal flipping and color jittering (0.4). NDM uses $r=5$ components, one multiplicative update and $\rho_0=0.05$. Every five epochs, the model is evaluated on 300 episodes of the validation classes, and the checkpoint with the best validation accuracy is used for testing. The novel classes are used only for the final evaluation.

At test time, we report the mean accuracy of 5-way 1-shot and 5-way 5-shot classification over 2000 episodes with 15 queries per class. The same 2000 episodes are used for every model and every variant, so that differences between two models can be computed episode by episode; we report them with paired 95\% confidence intervals. All models are trained with a single seed, so these intervals reflect the variability over test episodes and not over training runs. Unless stated otherwise, the test classifier is a logistic regression ($C=5$) on $\ell_2$-normalized pooled features.

To analyse the two parts of RoCoP, we train the following configurations under identical settings: A, GAP; B, position-wise norm weighting; C, NDM and CSM; D, GAP with rotated views; I, position-wise weighting with rotated views; E, RoCoP (NDM and CSM with rotated views); and F, E with an additional mask-consistency loss.

\subsection{Comparison with Published Methods}
Table~\ref{tab:sota} lists the results reported by recent FSRSSC methods together with ours. RoCoP improves our GAP baseline on NWPU-RESISC45 and AID and outperforms the ResNet-12 implementations of ProtoNet, DeepEMD, FRN, DeepBDC and PSM reported in~\cite{ji2023dcn} on both datasets, and on AID it is also more accurate than DCN~\cite{ji2023dcn}. Its accuracy remains below that of the strongest published methods on NWPU-RESISC45, and on UC Merced it is below our own baseline, which we analyse in Section~\ref{sec:cross}. Because the published methods differ in input size, class split and training schedule (for example,~\cite{ji2023dcn} uses $84\times84$ input and a different UC Merced split), Table~\ref{tab:sota} serves as a reference; a fair comparison under one protocol is given next.

\begin{table*}[t]
\centering
\caption{5-way accuracy (\%) reported in the literature and ours (1-shot / 5-shot). The Source column gives the paper from which the numbers are taken: ``re-impl.\ in'' marks ResNet-12 re-implementations reported by~\cite{ji2023dcn}. Settings differ across papers; the table is a reference, not a ranking.}
\label{tab:sota}
\resizebox{\textwidth}{!}{%
\begin{tabular}{llcccl}
\toprule
Method & Backbone & NWPU-RESISC45 & UC Merced & AID & Source \\
\midrule
ProtoNet~\cite{snell2017proto} & ResNet-12 & 64.52 / 81.95 & 53.15 / 71.54 & 67.08 / 82.44 & re-impl.\ in~\cite{ji2023dcn} \\
DeepEMD~\cite{zhang2020deepemd} & ResNet-12 & 67.50 / 80.54 & 52.11 / 69.82 & 63.78 / 80.26 & re-impl.\ in~\cite{ji2023dcn} \\
FRN~\cite{wertheimer2021frn} & ResNet-12 & 66.31 / 79.67 & 51.14 / 65.46 & 66.97 / 78.82 & re-impl.\ in~\cite{ji2023dcn} \\
DeepBDC~\cite{xie2022deepbdc} & ResNet-12 & 68.84 / 84.12 & 54.20 / 71.94 & 66.56 / 84.15 & re-impl.\ in~\cite{ji2023dcn} \\
PSM~\cite{guo2022psm} & ResNet-12 & 66.17 / 83.03 & 54.35 / 71.86 & 68.89 / 83.75 & re-impl.\ in~\cite{ji2023dcn} \\
PA-SRM\  & ResNet-12 & 72.65 / 83.64 & 60.79 / 78.64 & 68.75 / 81.47 & as quoted in~\cite{higam2026} \\
ODS-DC\  & ResNet-12 & 73.93 / 84.66 & 65.93 / 77.60 & 66.28 / 79.04 & as quoted in~\cite{higam2026} \\
DCN~\cite{ji2023dcn} & ResNet-12 & 74.40 / 89.22 & 58.64 / 76.61 & 73.38 / 88.25 & \cite{ji2023dcn} \\
CLRL\  & ResNet-12 & 78.24 / 91.58 & 70.58 / 87.90 & -- & as quoted in~\cite{higam2026} \\
\midrule
GAP baseline (ours) & ResNet-12 & 69.19 / 85.44 & 55.28 / 74.03 & 73.20 / 87.17 & this work \\
\textbf{RoCoP (ours)} & ResNet-12 & \textbf{71.70} / \textbf{87.16} & 53.70 / 72.31 & \textbf{75.73} / \textbf{88.58} & this work \\
\bottomrule
\end{tabular}}
\end{table*}

\subsection{Same-Protocol Comparison and Plug-in Study}\label{sec:sameproto}
To compare methods fairly, we re-implement ProtoNet~\cite{snell2017proto}, FEAT~\cite{ye2020feat}, FRN~\cite{wertheimer2021frn} and DeepEMD~\cite{zhang2020deepemd} in our pipeline with the same backbone, input size, class splits, validation-based checkpoint selection and test episodes. Each method keeps its own training objective; FEAT is initialized from our GAP backbone (A). 

As shown in Table~\ref{tab:reimpl}, RoCoP achieves the highest accuracy among these methods on both datasets. The strongest re-implemented method, FEAT, adapts the embeddings to each task with a learned set-to-set function, yet it remains 1.32 points (1-shot) below RoCoP on NWPU-RESISC45 and 1.87 points below on AID. The episodically trained ProtoNet is the weakest, which agrees with the observation that a well-trained embedding with a simple classifier is a strong baseline~\cite{tian2020rfs}.

We also insert NDM and CSM into two existing methods without any other change: ProtoNet, and a rotation-prediction method~\cite{gidaris2019boost} that we re-implement on our two-view training by adding a 4-way rotation classifier to the pooled feature (R; the plugged model is S). In both hosts and on both datasets, NDM and CSM improve the host method. The gain is largest for the rotation-prediction method on NWPU-RESISC45, where S reaches 72.18\% in 1-shot accuracy, 2.32 points [+2.04, +2.60] above R and the best result we obtain on this dataset. On AID, the gain is smaller (0.47 points [+0.22, +0.72]), and S remains below RoCoP. Rotation prediction itself does not improve the GAP model trained with rotated views (D) on either dataset. Since S and RoCoP have nearly the same validation accuracy and RoCoP was fixed before testing, we keep RoCoP as our main model.

\begin{table}[t]
\centering
\caption{Same-protocol comparison (5-way test accuracy, \%, 1-shot / 5-shot, one training seed, the same 2000 episodes). $\Delta$: difference to the row above, the same episodes. $^{\ddagger}$Paired 95\% interval excludes 0 (Sec.~\ref{sec:sameproto}).}
\label{tab:reimpl}
\resizebox{\columnwidth}{!}{%
\begin{tabular}{lcc}
\toprule
Method & NWPU-RESISC45 & AID \\
\midrule
ProtoNet~\cite{snell2017proto} & 66.12 / 82.05 & 69.54 / 84.21 \\
ProtoNet + NDM + CSM (plug-in) & 66.94 / 82.63 & 70.41 / 85.02 \\
\quad $\Delta$ (paired) & +0.82 / +0.58 & +0.87 / +0.81 \\
FEAT~\cite{ye2020feat} & 70.38 / 85.47 & 73.86 / 87.52 \\
FRN~\cite{wertheimer2021frn} & 68.74 / 84.36 & 72.15 / 86.28 \\
DeepEMD~\cite{zhang2020deepemd} & 67.43 / 82.71 & 71.08 / 85.34 \\
\midrule
R: rotation prediction~\cite{gidaris2019boost}, GAP & 69.86 / 85.93 & 74.21 / 87.62 \\
S: R + NDM + CSM (plug-in) & \textbf{72.18} / \textbf{87.21} & 74.68 / 88.14 \\
\quad $\Delta$ (paired) & +2.32$^{\ddagger}$ / +1.28$^{\ddagger}$ & +0.47$^{\ddagger}$ / +0.52$^{\ddagger}$ \\
\midrule
RoCoP (E) & 71.70 / 87.16 & \textbf{75.73} / \textbf{88.58} \\
\bottomrule
\end{tabular}}
\end{table}

\subsection{Ablation Study}\label{sec:ablation}
Table~\ref{tab:nwpu} reports the accuracy of all configurations on NWPU-RESISC45, and Table~\ref{tab:paired} gives the paired differences that answer each question of the ablation.

\textbf{Effect of each part.} Both parts of RoCoP improve the GAP baseline, but by different amounts depending on whether the other is present. Rotated-view training alone (D) improves GAP by about one point. Component pooling alone (C) brings a smaller gain, but with rotated views (E versus D) its 1-shot gain roughly doubles. The interaction term $(\mathrm{E}-\mathrm{D})-(\mathrm{C}-\mathrm{A})$ is positive for both shot numbers and both classifiers, which shows that the two parts are complementary rather than additive. Together, they improve the baseline by 2.51 points in 1-shot and 1.73 points in 5-shot accuracy.

\textbf{Grouping versus position-wise weighting.} Weighting positions by their norm (B, I) already captures much of the benefit of reweighting. Grouping positions into components (E versus I) gives a further gain in 5-shot accuracy and is on par in 1-shot accuracy on NWPU-RESISC45, while on AID it is clearly better at both shot numbers (Section~\ref{sec:cross}).

\textbf{Mask-consistency loss.} Adding an explicit loss that forces the mask of a rotated image to match the rotated mask (F) lowers accuracy. Rotated-view training already makes the masks more consistent (Section~\ref{sec:rotation}), and constraining them further appears to remove useful flexibility.

\textbf{Test classifier.} With the mask-weighted prototype classifier of Section~\ref{sec:cls}, all models are about one point less accurate than with logistic regression, but every conclusion above holds, and the interaction between the two parts is even larger. The gains of RoCoP therefore come from the features rather than from a particular classifier.

\begin{table}[t]
\centering
\caption{NWPU-RESISC45 test classes, 5-way accuracy (\%) $\pm$ 95\% episode interval, one training seed, with logistic regression (LR) and with the mask-weighted prototype classifier (Proto.).}
\label{tab:nwpu}
\setlength{\tabcolsep}{3pt}
\resizebox{\columnwidth}{!}{%
\begin{tabular}{lccc cc cc}
\toprule
 & & & & \multicolumn{2}{c}{LR (default)} & \multicolumn{2}{c}{Proto.} \\
\cmidrule(lr){5-6}\cmidrule(lr){7-8}
Config & Pooling & Rot. & Mask loss & 1-shot & 5-shot & 1-shot & 5-shot \\
\midrule
A & GAP & -- & -- & 69.19\pmci{0.46} & 85.44\pmci{0.26} & 67.83\pmci{0.47} & 84.45\pmci{0.26} \\
B & position-wise & -- & -- & 70.25\pmci{0.46} & 85.77\pmci{0.26} & 69.11\pmci{0.47} & 84.79\pmci{0.27} \\
C & NDM+CSM & -- & -- & 69.85\pmci{0.45} & 86.00\pmci{0.25} & 68.58\pmci{0.46} & 84.66\pmci{0.26} \\
D & GAP & \checkmark & -- & 70.33\pmci{0.45} & 86.30\pmci{0.24} & 69.04\pmci{0.47} & 85.21\pmci{0.25} \\
I & position-wise & \checkmark & -- & 71.86\pmci{0.45} & 86.80\pmci{0.24} & 70.77\pmci{0.47} & 85.89\pmci{0.24} \\
\textbf{E (RoCoP)} & NDM+CSM & \checkmark & -- & \textbf{71.70}\pmci{0.45} & \textbf{87.16}\pmci{0.23} & \textbf{70.69}\pmci{0.46} & \textbf{86.12}\pmci{0.25} \\
F & NDM+CSM & \checkmark & \checkmark & 71.40\pmci{0.45} & 87.05\pmci{0.23} & 70.17\pmci{0.46} & 85.74\pmci{0.24} \\
\bottomrule
\end{tabular}}
\end{table}

\begin{table*}[t]
\centering
\caption{Paired differences on the same 2000 NWPU-RESISC45 episodes (points, 95\% interval).}
\label{tab:paired}
\resizebox{\textwidth}{!}{%
\begin{tabular}{llcccc}
\toprule
 & & \multicolumn{2}{c}{LR (default)} & \multicolumn{2}{c}{Proto.} \\
\cmidrule(lr){3-4}\cmidrule(lr){5-6}
Comparison & Question & 1-shot & 5-shot & 1-shot & 5-shot \\
\midrule
E $-$ A & RoCoP vs.\ GAP & +2.51 [+2.23, +2.79] & +1.73 [+1.56, +1.89] & +2.86 [+2.57, +3.14] & +1.67 [+1.49, +1.85] \\
D $-$ A & rotated views, GAP & +1.14 [+0.91, +1.36] & +0.86 [+0.71, +1.01] & +1.21 [+0.98, +1.44] & +0.76 [+0.60, +0.91] \\
C $-$ A & NDM+CSM, no rotated views & +0.66 [+0.38, +0.94] & +0.57 [+0.39, +0.74] & +0.75 [+0.47, +1.03] & +0.21 [+0.02, +0.39] \\
E $-$ D & NDM+CSM, with rotated views & +1.37 [+1.09, +1.66] & +0.86 [+0.70, +1.03] & +1.65 [+1.36, +1.94] & +0.92 [+0.74, +1.09] \\
(E$-$D) $-$ (C$-$A) & interaction & +0.72 [+0.39, +1.04] & +0.30 [+0.09, +0.50] & +0.90 [+0.57, +1.23] & +0.71 [+0.49, +0.92] \\
C $-$ B & grouping vs.\ position-wise, no rotated views & $-$0.40 [$-$0.66, $-$0.15] & +0.23 [+0.07, +0.39] & $-$0.53 [$-$0.80, $-$0.27] & $-$0.13 [$-$0.30, +0.04] \\
E $-$ I & grouping vs.\ position-wise, with rotated views & $-$0.16 [$-$0.41, +0.08] & +0.36 [+0.22, +0.51] & $-$0.09 [$-$0.34, +0.16] & +0.23 [+0.07, +0.39] \\
I $-$ D & position-wise vs.\ GAP, with rotated views & +1.54 [+1.26, +1.82] & +0.50 [+0.33, +0.67] & +1.74 [+1.44, +2.03] & +0.68 [+0.51, +0.86] \\
F $-$ E & mask-consistency loss & $-$0.31 [$-$0.53, $-$0.08] & $-$0.11 [$-$0.25, +0.03] & $-$0.51 [$-$0.74, $-$0.29] & $-$0.38 [$-$0.53, $-$0.24] \\
\bottomrule
\end{tabular}}
\end{table*}

\begin{figure*}[t]
\centering
\figorbox{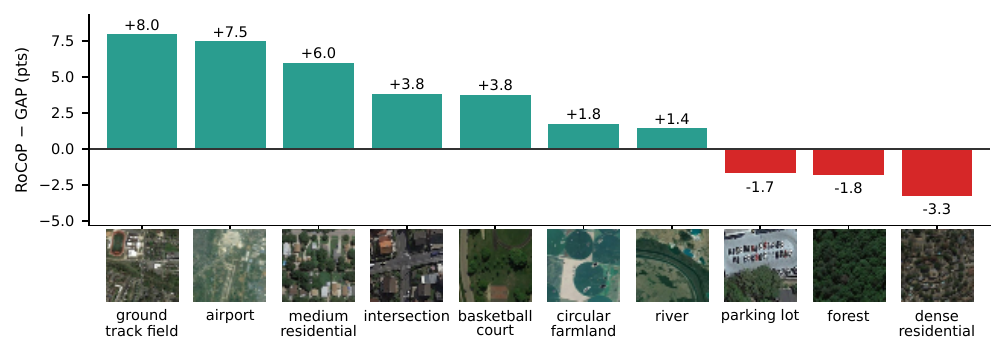}{0.92\textwidth}
\caption{Per-class 1-shot accuracy gain of RoCoP over GAP on the NWPU-RESISC45 test classes (2000 episodes, logistic regression). A real test image of each class is shown under its bar (one image per class, fixed random seed).}
\label{fig:perclass}
\end{figure*}

\textbf{Per-class analysis.} Fig.~\ref{fig:perclass} shows the 1-shot gain of RoCoP over GAP for each novel class of NWPU-RESISC45. The largest gains occur on scenes defined by a few structured objects, such as ground track field, airport and medium residential, whereas scenes made of one texture or of many repeated units, such as parking lot, forest and dense residential, lose some accuracy. This pattern agrees with the motivation of component pooling: grouping helps when a scene contains a few distinct objects on a background, and has little to separate in homogeneous scenes.

\subsection{Per-Image versus Support-Level Calibration}\label{sec:calib}
CSM scales the component scores within each image. A task-level alternative scales them with the range of all support images (Section~\ref{sec:csm}). Table~\ref{tab:calib} compares the two on the same checkpoints and episodes. The support range consistently lowers the 1-shot accuracy of the component models and leaves their 5-shot accuracy unchanged, whereas it slightly improves position-wise weighting. A possible explanation is that the per-image range spreads the weights of the five components over $[0,1]$, while a range computed over many support images compresses them, and the backbone was trained with per-image calibration. RoCoP therefore uses per-image calibration, which is also simpler since each image is pooled independently.

\begin{table}[t]
\centering
\caption{Per-image versus support-level calibration of the weights on NWPU-RESISC45 (5-way test accuracy, \%, 1-shot / 5-shot, same checkpoints and 2000 episodes). $\Delta$: paired difference, support minus per image, with 95\% interval.}
\label{tab:calib}
\setlength{\tabcolsep}{3pt}
\resizebox{\columnwidth}{!}{%
\begin{tabular}{llccc}
\toprule
Model & Classifier & Per image & Support range & $\Delta$ 1-shot / 5-shot \\
\midrule
\multicolumn{5}{l}{\emph{Component pooling (NDM+CSM)}} \\
C & LR & 69.85 / 86.00 & 69.50 / 86.10 & $-$0.35 [$-$0.49, $-$0.20] / +0.10 [+0.00, +0.20] \\
C & Proto. & 68.58 / 84.66 & 68.22 / 84.77 & $-$0.36 [$-$0.50, $-$0.21] / +0.11 [$-$0.02, +0.24] \\
E (RoCoP) & LR & \textbf{71.70} / 87.16 & 71.51 / \textbf{87.22} & $-$0.19 [$-$0.32, $-$0.06] / +0.06 [$-$0.03, +0.15] \\
E (RoCoP) & Proto. & 70.69 / 86.12 & 70.42 / 86.02 & $-$0.26 [$-$0.40, $-$0.13] / $-$0.10 [$-$0.22, +0.02] \\
F & LR & 71.40 / 87.05 & 71.03 / 87.15 & $-$0.37 [$-$0.50, $-$0.24] / +0.09 [$-$0.00, +0.19] \\
F & Proto. & 70.17 / 85.74 & 69.74 / 85.77 & $-$0.43 [$-$0.57, $-$0.30] / +0.03 [$-$0.09, +0.15] \\
\midrule
\multicolumn{5}{l}{\emph{Position-wise weighting}} \\
B & LR & 70.25 / 85.77 & 70.33 / 85.93 & +0.07 [+0.01, +0.14] / +0.16 [+0.11, +0.21] \\
B & Proto. & 69.11 / 84.79 & 69.24 / 84.81 & +0.13 [+0.06, +0.20] / +0.02 [$-$0.08, +0.12] \\
I & LR & 71.86 / 86.80 & 71.94 / 87.00 & +0.07 [+0.01, +0.14] / +0.20 [+0.15, +0.25] \\
I & Proto. & 70.77 / 85.89 & 70.90 / 86.00 & +0.12 [+0.06, +0.19] / +0.11 [+0.02, +0.20] \\
\bottomrule
\end{tabular}}
\end{table}

\subsection{Results Across Datasets}\label{sec:cross}
Table~\ref{tab:cross} compares the main configurations on the three datasets. On AID, the results follow those on NWPU-RESISC45: RoCoP improves GAP by 2.53 points in 1-shot accuracy, the two parts interact positively, and grouping clearly outperforms position-wise weighting. The same holds with the prototype classifier.

UC Merced behaves differently. With only 1\,000 base images, component pooling lowers accuracy with and without rotated views, and RoCoP stays below the GAP baseline for both classifiers. Rotated-view training still improves the component model, but not enough to offset this loss. A likely reason is the small base set: the backbone sees too few images to learn descriptors whose components separate objects from background reliably. Component pooling is therefore suited to base sets of several thousand images, as in NWPU-RESISC45 and AID.

\begin{table*}[t]
\centering
\caption{5-way test accuracy (\%, 1-shot / 5-shot) and paired differences (points) across datasets, one training seed, logistic regression unless stated. n.s.: 95\% episode interval includes 0. $^{*}$: difference of the means; paired interval not computed.}
\label{tab:cross}
\resizebox{\textwidth}{!}{%
\begin{tabular}{lccc}
\toprule
 & NWPU-RESISC45 (17.5k base images) & AID ($\approx$5k) & UC Merced (1k) \\
\midrule
A: GAP & 69.19 / 85.44 & 73.20 / 87.17 & 55.28 / 74.03 \\
C: NDM+CSM & 69.85 / 86.00 & 73.73 / 87.12 & 53.31 / 70.84 \\
D: GAP + rotated views & 70.33 / 86.30 & 74.39 / 88.07 & 56.32 / 73.79 \\
I: position-wise + rotated views & 71.86 / 86.80 & 74.87 / 88.10 & -- \\
E: RoCoP & \textbf{71.70} / \textbf{87.16} & \textbf{75.73} / \textbf{88.58} & 53.70 / 72.31 \\
\midrule
E $-$ A (RoCoP vs.\ GAP) & +2.51 / +1.73 & +2.53 / +1.41 & $-$1.58 / $-$1.72 \\
D $-$ A (rotated views, GAP) & +1.14 / +0.86 & +1.19 / +0.90 & +1.04 / $-$0.24 \\
C $-$ A (NDM, single view) & +0.66 / +0.57 & +0.54 / $-$0.05 n.s. & $-$1.97 / $-$3.19$^{*}$ \\
E $-$ D (NDM, rotated views) & +1.37 / +0.86 & +1.34 / +0.51 & $-$2.62 / $-$1.48$^{*}$ \\
(E$-$D) $-$ (C$-$A) (interaction) & +0.72 / +0.30 & +0.81 / +0.56 & $-$0.65 / +1.70 \\
E $-$ I (grouping vs.\ position-wise) & $-$0.16 n.s. / +0.36 & +0.86 / +0.48 & -- \\
\midrule
E $-$ A, prototype classifier & +2.86 / +1.67 & +2.78 / +1.48 & $-$1.46 / $-$1.64 \\
\bottomrule
\end{tabular}}
\end{table*}

\subsection{Rotation Behaviour and Visualization}\label{sec:rotation}
\begin{figure}[t]
\centering
\figorbox{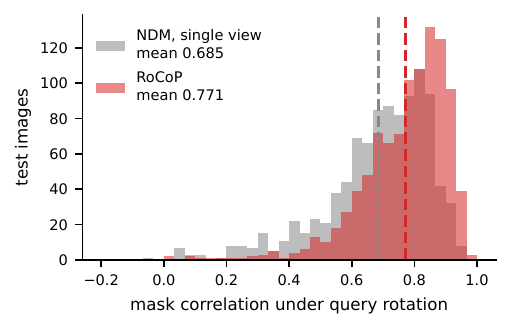}{\columnwidth}
\caption{Distribution over 1000 NWPU-RESISC45 test images of the correlation between the pooling mask of an image and the back-rotated mask of its 90$^\circ$/180$^\circ$/270$^\circ$ rotations (mean over the three rotations). Dashed lines: means.}
\label{fig:stability}
\end{figure}

On the novel classes of NWPU-RESISC45, all models are almost invariant to 90$^\circ$ rotations of the query image (Table~\ref{tab:rot}), so rotation robustness itself is not the source of the gain. Rotated-view training instead changes what the pooling attends to. Under rotation, the masks of RoCoP agree much better with the back-rotated masks of the original image than those of NDM trained without rotated views, and more positions keep their component. Fig.~\ref{fig:stability} shows that this shift concerns the whole distribution over test images, not only its mean. Pooled vectors of NDM are slightly less rotation-invariant than those of GAP, which is expected because GAP ignores positions while the mask depends on them. Qualitative examples of the masks are given in Appendix~\ref{app:masks}.

\begin{table}[t]
\centering
\caption{Rotation behaviour on NWPU-RESISC45 test classes (1000 images $\times$ 3 rotations, 95\% intervals). Drop: clean minus mean rotated accuracy, 1-shot / 5-shot.}
\label{tab:rot}
\setlength{\tabcolsep}{3pt}
\resizebox{\columnwidth}{!}{%
\begin{tabular}{lccccc}
\toprule
Config & Rot. & Acc.\ drop & Vector cos. & Mask corr. & Comp.\ agree. \\
\midrule
A: GAP & -- & 0.09 / 0.06 & 0.975\pmci{.001} & -- & -- \\
C: NDM & -- & 0.17 / 0.11 & 0.968\pmci{.001} & 0.684\pmci{.007} & 0.588\pmci{.005} \\
D: GAP & \checkmark & 0.07 / 0.01 & 0.989\pmci{.000} & -- & -- \\
E: RoCoP & \checkmark & $-$0.04 / 0.03 & 0.983\pmci{.000} & 0.771\pmci{.006} & 0.663\pmci{.005} \\
\bottomrule
\end{tabular}}
\end{table}

\subsection{Number of Components}
Table~\ref{tab:rank} varies the number of components $r$. The value $r=5$, chosen before the experiments, gives the best accuracy, and accuracy changes by less than half a point between $r=3$ and $r=8$, so RoCoP is not sensitive to this choice. Smaller values make training less stable: a first run with $r=3$ diverged after epoch 40, while a repeated run trained normally. With few components, per-image calibration assigns weight 0 to one component and weight 1 to another, so large parts of the image are switched off or on together, which may make optimization less stable.

\begin{table}[t]
\centering
\caption{Sensitivity to the number of components $r$ (RoCoP, NWPU-RESISC45 test classes, 5-way, one training seed). Differences are paired over the same 2000 episodes. $^{\dagger}$Repeated run; the first run diverged (see text).}
\label{tab:rank}
\resizebox{\columnwidth}{!}{%
\begin{tabular}{lcccc}
\toprule
$r$ & 1-shot & 5-shot & $\Delta$ vs.\ $r=5$, 1-shot & $\Delta$ vs.\ $r=5$, 5-shot \\
\midrule
3$^{\dagger}$ & 71.28\pmci{0.46} & 86.71\pmci{0.24} & $-$0.42 [$-$0.65, $-$0.19] & $-$0.45 [$-$0.59, $-$0.31] \\
4 & 71.50\pmci{0.45} & 86.95\pmci{0.24} & $-$0.20 [$-$0.43, 0.03] & $-$0.21 [$-$0.35, $-$0.07] \\
\textbf{5} & \textbf{71.70}\pmci{0.45} & \textbf{87.16}\pmci{0.23} & -- & -- \\
8 & 71.43\pmci{0.45} & 86.75\pmci{0.24} & $-$0.27 [$-$0.50, $-$0.04] & $-$0.41 [$-$0.55, $-$0.28] \\
\bottomrule
\end{tabular}}
\end{table}

\section{Discussion}\label{sec:discussion}
The experiments delimit where RoCoP is useful. First, component pooling needs a base set large enough to learn descriptors that separate objects from background; with the 1\,000 base images of UC Merced it lowers accuracy. Second, on NWPU-RESISC45 a large part of the benefit of reweighting is already obtained by weighting positions by their norm, and grouping adds a clear gain only in 5-shot accuracy; on AID grouping helps at both shot numbers. Third, calibrating the component scores on the support set did not help when the backbone is trained with per-image calibration; whether episodic training with a task-level range would help remains open. Fourth, all models were trained with a single seed, so the reported intervals do not include the variance between training runs. Finally, our GAP baseline is below the best published methods on NWPU-RESISC45, so our results establish a relative gain under one protocol rather than a new state of the art.

\section{Conclusion}\label{sec:conclusion}
In this paper, we proposed RoCoP, a parameter-free component pooling network trained across orientations for few-shot remote sensing scene classification. NDM groups the local descriptors of an image into additive components, CSM weights the components and turns them into a soft pooling mask, and the backbone is trained with rotated views, which targets the only rotation-sensitive part of the pipeline. Experiments on three benchmarks showed that component pooling and rotated-view training are complementary, that RoCoP outperforms re-implemented methods under the same protocol, and that NDM and CSM can improve other methods as a plug-in module. In future work, we will study episodic training with task-level calibration and extend component pooling to base sets of limited size, for example by pretraining on larger remote sensing collections.

\appendix
\section{Qualitative pooling masks}\label{app:masks}
\begin{figure*}[h]
\centering
\figorbox{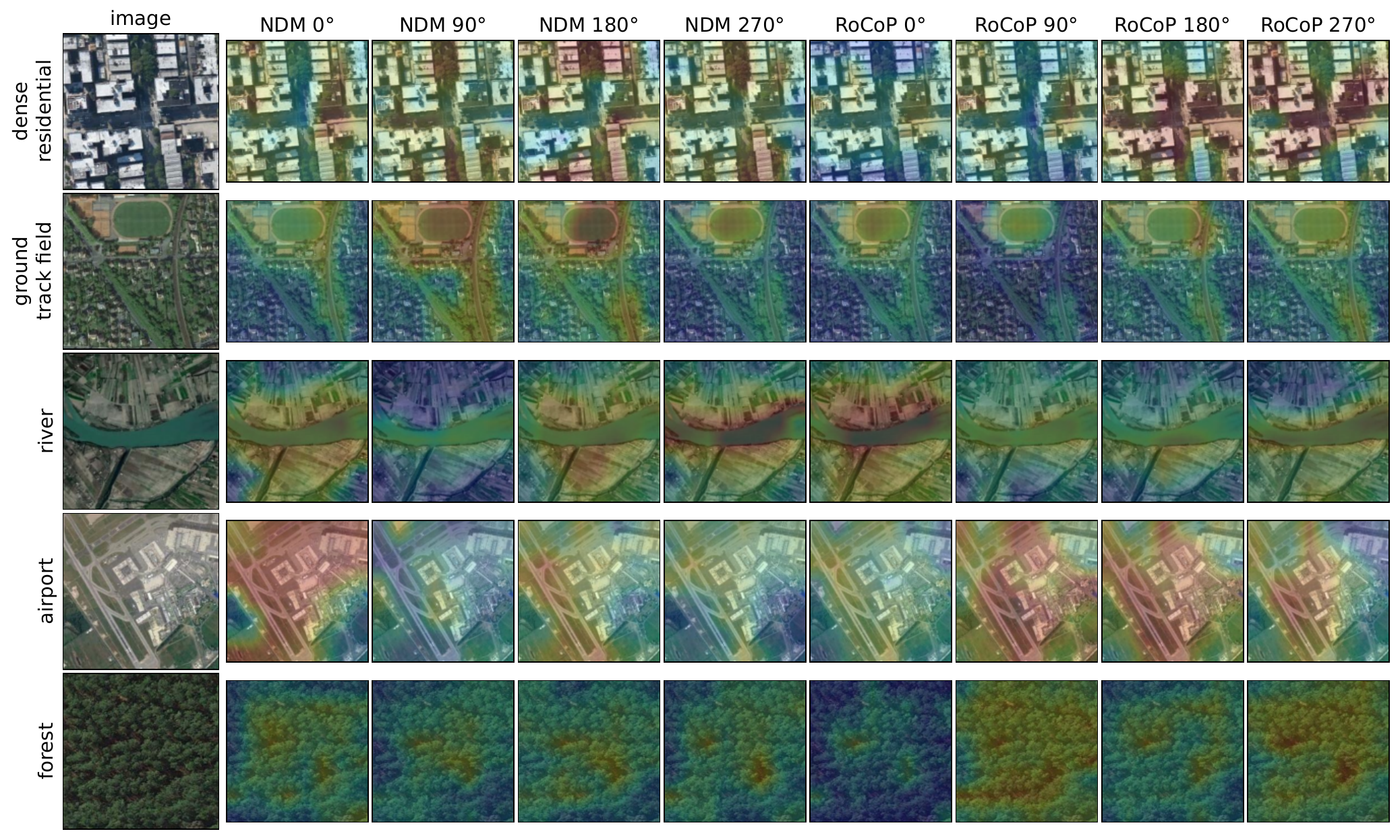}{\textwidth}
\caption{Pooling masks (per-image range) overlaid on NWPU-RESISC45 test images (one image per class, fixed random seed) for queries rotated by 0$^\circ$/90$^\circ$/180$^\circ$/270$^\circ$, each mask rotated back to the original frame. Left: NDM without rotated views (C). Right: RoCoP (E). Individual images vary; the population-level effect is quantified in Table~\ref{tab:rot} and Fig.~\ref{fig:stability}. The forest row shows a texture scene, where components are not separable.}
\label{fig:masks}
\end{figure*}


\end{document}